\documentclass[conference]{IEEEtran}
\IEEEoverridecommandlockouts
\usepackage{amsmath, amssymb, amsfonts}

\usepackage{multirow}
\usepackage{array}
\usepackage{multicol}
\usepackage{makecell}
\usepackage{colortbl}
\usepackage[most]{tcolorbox}
\usepackage{tabularx}
\usepackage{cite}
\usepackage{amsmath,amssymb,amsfonts}
\usepackage{algorithmic}
\usepackage{graphicx}
\usepackage{textcomp}
\usepackage{xcolor}
\def\BibTeX{{\rm B\kern-.05em{\sc i\kern-.025em b}\kern-.08em
    T\kern-.1667em\lower.7ex\hbox{E}\kern-.125emX}}

\usepackage{cuted}
\usepackage{capt-of}
\usepackage{indentfirst}
\usepackage{makecell} 
\usepackage{booktabs}
\usepackage{caption}
\definecolor{darkgreen}{RGB}{0,90,50}
\usepackage{subcaption}
\usepackage{hyperref}
\usepackage{stfloats}

\usepackage{eso-pic} 

\hypersetup{
    colorlinks=true,
    linkcolor=blue,   
    citecolor=red,   
    urlcolor=darkgreen     
}

\IEEEoverridecommandlockouts

\begin{document}

\title{SkillLens: Action-Guided Dense Visual Dynamics for Proficiency Estimation}

\author{
    Deokhyun Ahn\textsuperscript{1,3,$\dagger$}, 
    Hyukjin Kim\textsuperscript{2,3,$\dagger$}, 
    Jae-Ho Han\textsuperscript{1},
    Bumsub Ham\textsuperscript{2},
    Heeseung Choi\textsuperscript{3}, 
    Ig-Jae Kim\textsuperscript{3},
    Haksub Kim\textsuperscript{3,*}
    \\
    \vspace{0.2cm} 
    \textsuperscript{1}Korea University
    \textsuperscript{2}Yonsei University
    \textsuperscript{3}Korea Institute of Science and Technology (KIST)
}

\IEEEaftertitletext{%
    \vspace{-2.5em}
    \noindent
    \begin{minipage}{\textwidth}
        \centering
        \includegraphics[
            width=0.96\linewidth
        ]{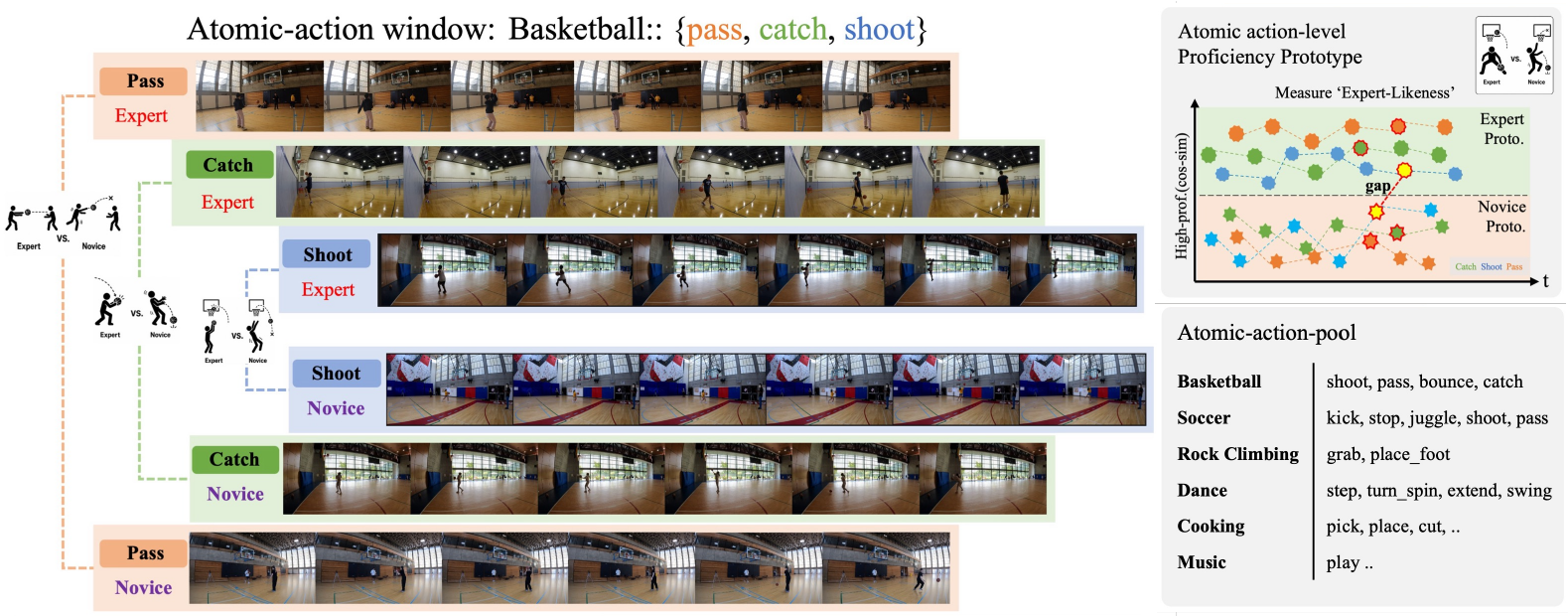}
        \captionof{figure}{
            \textbf{Where should we look to assess skill?}
            We use atomic-action windows to compare
            dense visual dynamics against expert and
            novice prototypes, yielding an action-level
            proficiency margin.
        }
        \label{fig_teaser}
    \end{minipage}
}

\maketitle



\maketitle

\AddToShipoutPictureFG*{%
  \AtPageLowerLeft{%
    \put(\LenToUnit{0.75in},\LenToUnit{0.55in}){%
      \parbox[b]{0.46\textwidth}{%
        \rule{0.4\columnwidth}{0.4pt}\\[4pt]
        \fontsize{8}{9}\selectfont
        $\dagger$ These authors contributed equally to this work.\\
        $^{*}$ Corresponding author.
      }%
    }%
  }%
}

\begin{abstract}
Human proficiency estimation is challenging because
skill is a high-level and often subjective property that is not
captured by action identity alone. Existing approaches leverage
multimodal observations such as ego-exo video, gaze, pose, and
inertial signals, but the gap between sensor-level evidence and
abstract proficiency labels remains difficult to bridge. 
We present \textit{SkillLens}, a proficiency estimation framework that uses
atomic action descriptions as temporal guidance, rather than direct semantic
supervision, to localize proficiency-relevant windows for dense visual-dynamics
extraction.
Each window is summarized into an action-dynamics feature and scored
by a high-low prototype margin against higher- and lower-
proficiency executions from the same activity-action category.
Fusing these ego/exo PDM cues with a global ego-exo-gaze
baseline through CORAL ordinal calibration, our method achieves
0.542 accuracy on Ego-Exo4D.
Our project page is available at \url{https://skill-lens.net/}.

\end{abstract}

\section{Introduction}

Proficiency estimation remains difficult even when ego video, exo video, and gaze
are available. The challenge is not simply to recognize which action is being
performed, but to judge how well it is executed~\cite{parmar2019what,pan2025basket}.
Recent benchmarks also highlight the difficulty of fine-grained, expert-level
understanding of skilled activities~\cite{yi2025exact}. Prior work has explored
fine-grained ego-exo representations and gaze-action relationships
\cite{xue2023viewinvariant,li2023beholder,mazzamuto2025gazing}.
A video-gaze baseline can capture global activity context, but sparse
full-take sampling may dilute the brief execution moments where proficiency
differences emerge~\cite{doughty2019pros,xu2022finediving}.
Studies of procedural activities have described human behavior through
steps and fine-grained action segments
\cite{song2023goalstep,perrett2025hdepic,gong2023activity}.
As illustrated in Fig.~\ref{fig_teaser}, we use atomic action descriptions as
temporal guidance to identify where proficiency-relevant execution cues occur.
An atomic action description is a short language annotation associated with a
local action segment, such as \textit{shoot}, \textit{pass}, or \textit{cut}.
We do not use these descriptions as direct semantic supervision; instead, we use
them to localize action-specific windows and extract dense visual dynamics.

Prior work has investigated skill-relevant temporal attention
\cite{doughty2019pros}, fine-grained and segment-level action quality assessment
\cite{xu2022finediving,mazzamuto2026proskill,joung2023action},
reference-based comparisons of execution quality~\cite{yu2021groupaware},
and skill-attribute representations~\cite{ashutosh2025skillattributes}.
Building on these directions, we structure proficiency estimation around where
execution quality should be measured, which high- and low-proficiency references
it should be compared against, and how predictions should be calibrated under
ordered skill levels. 
We summarize our contributions as follows:
\begin{itemize}
    \item \textit{Activity-aware aspect expert routing.}
    We formulate proficiency estimation as activity-dependent evidence
    aggregation. Ego, exo, and gaze features are organized into aspect-oriented
    expert branches aligned with modality-specific evidence, such as gaze-related
    signals from Aria eye-gaze and ego video or interaction-related cues from
    ego-centric visual features. An activity-aware router combines a fixed
    activity-aspect prior with learned sample-conditioned gates to re-weight expert
    evidence for each~input.

    \item \textit{Atomic-action-guided high-low prototype margin.}
    We use atomic action descriptions to define where execution quality should
    be measured. Each localized action window is summarized into an
    action-dynamics feature and compared with high- and low-proficiency
    prototypes from the same activity-action category, producing relative
    evidence of expert-like versus novice-like execution.

    \item \textit{Ordinal proficiency calibration.}
    We treat proficiency labels as ordered levels rather than independent
    classes. Activity-normalized prototype margins are fused with the global
    baseline prediction through a CORAL~\cite{coral} ordinal head, enabling
    calibration under the natural Novice-to-Late-Expert progression.
\end{itemize}

\begin{figure*}[t!]
\centering
\includegraphics[width=1\textwidth]{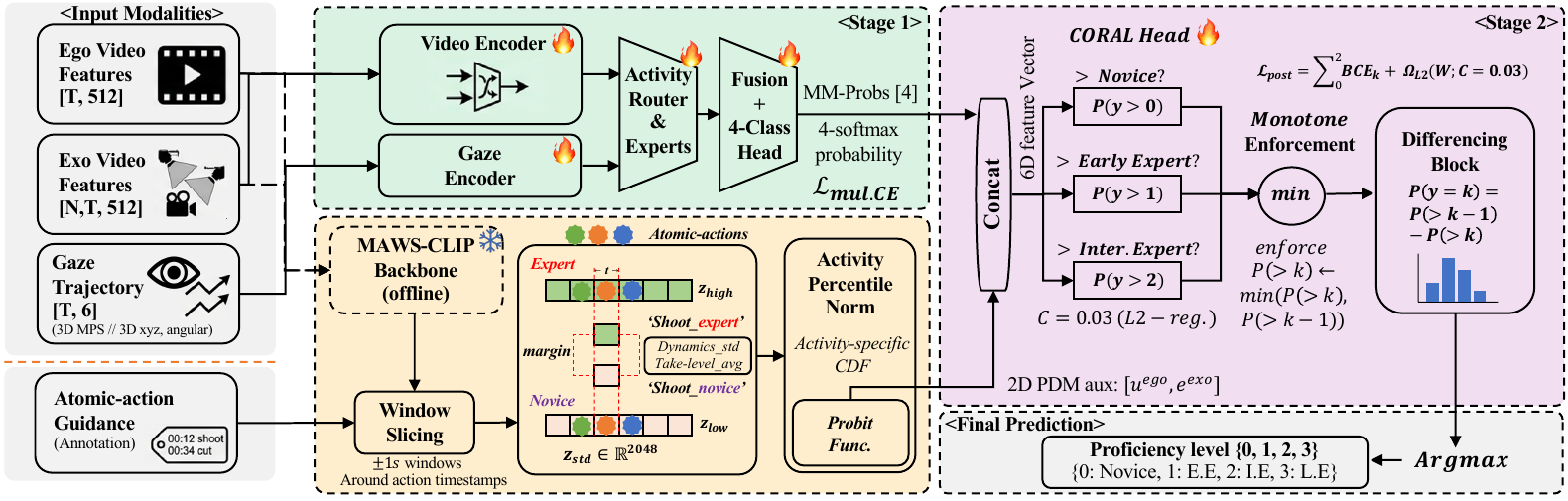} 
\caption{\textbf{Overview of \textbf{SkillLens}.} A frozen video-gaze classifier provides multimodal class probabilities, while atomic-action guidance localizes dense MAWS windows to compute an activity-normalized proficiency dynamics score. We concatenate both signals and use a CORAL ordinal head to predict the final proficiency level.}
\label{fig:overview}
\vspace{-1em}
\end{figure*}
\section{Methods}

\subsection{Multi-modal Baseline with Activity-Aware Aspect Routing}

Before calibration, we train a Stage-1 multi-modal baseline that maps ego, exo,
and gaze features into a 4D proficiency probability vector. Let
$H^m \in \mathbb{R}^{T \times d}$ denote the temporal feature sequence of
modality $m \in \{\mathrm{ego}, \mathrm{exo}, \mathrm{gaze}\}$, and let
$\mathcal{A}$ denote a set of predefined aspect-oriented expert branches.
These branches organize modality-specific evidence around activity-relevant
factors, such as gaze-related evidence from Aria eye-gaze and ego video,
interaction-related evidence from ego-centric visual features, and
timing/procedure-related evidence from ego-exo video context. They are not
intended to explicitly extract semantic labels; instead, they provide an
inductive structure for aggregating modality-specific evidence.

For each aspect branch $a \in \mathcal{A}$, an expert receives a designated
subset of modality features $\mathcal{S}_a$ and produces an aspect-oriented
evidence vector $e_a$ and an auxiliary aspect-polarity prediction:
\[
e_a =
\mathrm{AttnPool}
\left(
\phi_a\left([H^m]_{m \in \mathcal{S}_a}\right)
\right),
\qquad
p_a = \mathrm{softmax}(W_a e_a),
\]
where $\phi_a$ is an aspect-specific MLP and $p_a \in \mathbb{R}^{3}$ represents
novice-like, unknown, and expert-like polarity.

The activity-aware router re-weights aspect expert evidence by combining a
fixed activity-aspect relevance prior with learned sample-conditioned gates.
Given the activity identity and the global multi-modal context $c$, the router
computes
\[
w =
\mathrm{softmax}
\left(
f_{\mathrm{rtr}}([E(\alpha);c]) + \log(\pi_{\alpha}+\epsilon)
\right),
\]
where $E(\alpha)$ is the activity embedding, $\pi_{\alpha}$ is a fixed
activity-aspect relevance prior, and $\epsilon$ avoids numerical instability.
The prior provides an activity-level bias, while the learned gate
$f_{\mathrm{rtr}}$ adapts the expert weights for each input sample.
The router-weighted evidence is then fused into the baseline prediction:
\[
\bar{e} = \sum_{a \in \mathcal{A}} w_a e_a,
\qquad
p_{\mathrm{base}} = \mathrm{softmax}(W_o \bar{e}) \in \mathbb{R}^{4}.
\]

The Stage-1 baseline is trained with a multi-task classification loss:
\[
\mathcal{L}_{\mathrm{mul.CE}}
=
\mathcal{L}_{\mathrm{cls}}
+
\lambda_{\mathrm{asp}}\mathcal{L}_{\mathrm{asp}}
+
\lambda_{\mathrm{rtr}}\mathcal{L}_{\mathrm{rtr}}
+
\lambda_{\mathrm{con}}\mathcal{L}_{\mathrm{con}}.
\]
Here, $\mathcal{L}_{\mathrm{cls}}$ is the overall proficiency classification
loss, $\mathcal{L}_{\mathrm{asp}}$ weakly supervises aspect-polarity predictions
when commentary-derived pseudo-labels are available, $\mathcal{L}_{\mathrm{rtr}}$
regularizes the routing weights, and $\mathcal{L}_{\mathrm{con}}$ encourages
consistency between aspect-level evidence and the overall proficiency
prediction. Commentary-derived aspect labels are used only during training and
are not required at inference.

\subsection{Dense Action Windows and High-Low Prototype Dynamics Margin (PDM)}

The Stage-1 baseline produces a global proficiency estimate, but sparse video
features can miss short execution dynamics that distinguish proficiency levels.
We therefore extract dense local features around atomic action timestamps.
For each stream $s \in \{\mathrm{ego}, \mathrm{exo}\}$ and timestamp $t$, we
extract dense MAWS features
$X_t^{(s)}=\{x_i^{(s)}\}_{i=1}^{N}$ within a bounded
$[t-1\,\mathrm{s},t+1\,\mathrm{s}]$ window, where $N\approx 60$ frames at 30 fps.
We summarize local visual dispersion by the temporal standard deviation
feature
\[
z_t^{(s)}
=
\sqrt{
\frac{1}{N}
\sum_{i=1}^{N}
\left(x_i^{(s)}-\bar{x}_t^{(s)}\right)^2
},
\]
where $\bar{x}_t^{(s)}$ is the mean feature within the window and the square and
square root are applied element-wise. 
This action-dynamics feature captures the temporal variation of dense visual
features within a local action window without requiring additional frame-level supervision.

Measuring raw similarity to expert prototypes can be biased by
activity-specific visual context shared across proficiency levels.
Thus, we use a relative high-low prototype margin.
Using only the train set, we aggregate action-dynamics features into activity-action buckets
$b=(\alpha,\beta)$, where $\alpha$ is the activity and $\beta$ is the atomic
action type, such as \textit{Basketball:shoot}.
For each stream $s$ and bucket $b$, we define a high prototype
$\mu_{\mathrm{high}}^{(s,b)}$ as the mean feature of Intermediate/Late experts,
and a low prototype $\mu_{\mathrm{low}}^{(s,b)}$ as the mean feature of
Novice/Early performers.
The collection of high-low prototypes over all activity-action buckets forms
the train-time prototype bank used for PDM scoring.

For a test action-dynamics feature $z_t^{(s)}$, the window-level prototype dynamics margin is
\[
m_t^{(s)}
=
\cos\left(z_t^{(s)},\mu_{\mathrm{high}}^{(s,b)}\right)
-
\cos\left(z_t^{(s)},\mu_{\mathrm{low}}^{(s,b)}\right).
\]
A positive value indicates greater similarity to high-proficiency executions
within the same activity-action bucket.
For each take and stream, we average the window-level margins over all available
atomic-action windows to obtain a take-level stream-specific PDM score
$m^{(s)}$.

In the main action-annotated setting, $z_t^{(s)}$ is extracted from windows
centered at atomic-action timestamps.
The same PDM computation can also be applied to \textit{annotation-free}
temporal proposals by scoring each proposal against all train-time high-low
prototypes associated with the same activity and taking the maximum margin
before activity-level normalization.

\subsection{Ordinal Calibration via CORAL Head}

Since motion scales vary across activities, we normalize each take-level
stream-specific PDM score using the empirical CDF of the train set within the
same activity:
\[
u^{(s)}
=
\Phi^{-1}
\left(
F_{\alpha}^{\mathrm{train}}
\left(
m^{(s)}
\right)
\right),
\qquad
s\in\{\mathrm{ego},\mathrm{exo}\}.
\]
We concatenate the normalized ego/exo PDM scores with the Stage-1 baseline
probability vector:
\[
v =
[p_{\mathrm{base}};\,u^{(\mathrm{ego})};\,u^{(\mathrm{exo})}].
\]
Missing exo PDM values are zero-filled in our experiments.

To leverage the inherent order of proficiency labels, we fit a Stage-2 CORAL
ordinal head on the frozen features. This reformulates the 4-class problem into
three cumulative binary boundaries,
\[
q_k = P(y>k \mid v), \qquad k\in\{0,1,2\}.
\]
The post-hoc objective is
\[
\mathcal{L}_{\mathrm{post}}
=
\sum_{k=0}^{2}
\mathrm{BCE}
\left(
\mathbf{1}\{y>k\}, q_k
\right)
+
\Omega_{\mathrm{L2}}(W; C=0.03),
\]
where $\Omega_{\mathrm{L2}}$ denotes the L2 penalty on the logistic regression
weights with $C=0.03$. We implement the CORAL head as an L2-regularized logistic regression
head with $C=0.03$.
After enforcing $q_0 \ge q_1 \ge q_2$, the final class probabilities are
obtained by differencing the cumulative probabilities.\\

\section{Experiments}

\subsection{Experimental Setup}

\noindent\textbf{Dataset and metrics.}
We evaluate our method on the Ego-Exo4D\cite{grauman2024ego} proficiency estimation task following the standard validation protocol~\cite{bianchi2026profvlm, bianchi2026skillformer, wu2025skillsight}.
The model predicts four ordered proficiency levels: Novice, Early Expert, Intermediate Expert, and Late Expert.
While prior work mainly reports accuracy, accuracy alone can obscure class imbalance and ordinal prediction errors.
We therefore additionally report macro-F1 for class-balanced performance, MAE for ordinal error, and Spearman correlation for ranking consistency.

\noindent\textbf{Implementation details.}
We use frozen 16-frame ego/exo video features and gaze trajectories, with temporal video Transformers, a GRU-based gaze encoder, and activity-aware aspect routing.
Stage-1 training uses AdamW ($\mathrm{lr}=2{\times}10^{-4}$, weight decay $10^{-4}$), batch size 8, gradient clipping 1.0, mixed precision, and $(\lambda_{\mathrm{asp}},\lambda_{\mathrm{rtr}},\lambda_{\mathrm{con}})=(0.5,0.01,0.1)$.
The CORAL head is fit on frozen $[p_{\mathrm{base}};u^{(\mathrm{ego})};u^{(\mathrm{exo})}]$ features with $C=0.03$, and all results are averaged over three seeds.

\begin{table}[h]
\centering
\caption{Proficiency benchmark results on Ego-Exo4D.}
\label{tab:quantitative_results}
\resizebox{\columnwidth}{!}{%
\begin{tabular}{lccccc}
\toprule
Method & Pretrained & Acc & mF1 & MAE & Spearman \\
\midrule
TimeSFormer\cite{bertasius2021space} & HowTo100M & 0.468 & - & - & - \\
EgoPulseFormer\cite{braun2025egoppg} & EgoPPG-DB & 0.453 & - & - & - \\
SkillFormer\cite{bianchi2026skillformer} & K600 & 0.459 & - & - & - \\
ProfVLM (AGP)\cite{bianchi2026profvlm} & K600 + SmoLM2 & 0.442 & - & - & - \\
SkillSight\cite{wu2025skillsight} & EgoVLPv2 & 0.501 & - & - & - \\
\midrule
\rowcolor{gray!18}
\textbf{SkillLens} & MAWS-CLIP-2B & \textbf{0.542} & \textbf{0.517} & \textbf{0.654} & \textbf{0.468} \\
\bottomrule
\end{tabular}%
}
\end{table}

\subsection{Results and Analysis}

\noindent\textbf{Main benchmark.}
Table~\ref{tab:quantitative_results} compares the proposed model with prior Ego-Exo4D proficiency estimation baselines.
Our full model achieves the best accuracy among the compared methods.
Because most previous methods do not report macro-F1, MAE, or Spearman correlation, we report these additional metrics for our model to better reflect class balance, ordinal error, and ranking consistency.

\begin{table}[t] 
\centering
\caption{Per-proficiency level performance comparison.}
\label{tab:performance_comparison} 
\resizebox{\columnwidth}{!}{%
\begin{tabular}{lcccc} 
\toprule
 & \multicolumn{2}{c}{F1} & \multicolumn{2}{c}{Precision} \\ 
\midrule
Class & Baseline & Proposed & Baseline & Proposed \\
\midrule
Novice & 0.295 & \textbf{0.418} & 0.275 & \textbf{0.341} \\
Early Expert & 0.524 & \textbf{0.582} & 0.514 & \textbf{0.550} \\
Intermediate Expert & 0.423 & \textbf{0.430} & 0.511 & \textbf{0.553} \\
Late Expert & 0.494 & \textbf{0.537} & 0.495 & \textbf{0.536} \\
\bottomrule
\end{tabular}%
}

\end{table}

\begin{table}[t!]
\centering
\caption{Per-activity performance comparison.}
\label{tab:activity_acc_single}
\resizebox{\columnwidth}{!}{%
\begin{tabular}{lcccccc}
\toprule
\textbf{Activity} & \textbf{SkillLens} & \textbf{w/o gaze} & \textbf{w/o PDM} & \textbf{w/o CORAL} & \textbf{w/o router} & \textbf{w/o aspect} \\
\midrule
Basketball & 0.659 & 0.565 & 0.585 & 0.630 & 0.711 & 0.585 \\
Cooking & 0.433 & 0.467 & 0.517 & 0.500 & 0.350 & 0.433 \\
Dance & 0.572 & 0.501 & 0.580 & 0.570 & 0.549 & 0.493 \\
Music & 0.685 & 0.556 & 0.778 & 0.741 & 0.481 & 0.537 \\
Rock Climbing & 0.439 & 0.408 & 0.357 & 0.366 & 0.399 & 0.441 \\
Soccer & 0.576 & 0.636 & 0.606 & 0.606 & 0.636 & 0.606 \\
\bottomrule
\end{tabular}%
}
\vspace{-1em}
\end{table}
\begin{table}[t]
\centering
\caption{Ablation on the Ego-Exo4D, with 3-seed mean.}
\label{tab:ablation_ladder}
\resizebox{\columnwidth}{!}{%
\begin{tabular}{llcccccccccc}
\toprule
& Stage & V & G & Rtr & Asp & PDM & CORAL & Acc & mF1 & MAE & Sp \\
\midrule
1 & Baseline (video only)
& \checkmark & -- & -- & -- & -- & --
& 0.430 & 0.383 & 0.796 & 0.382 \\

2 & + gaze
& \checkmark & \checkmark & -- & -- & -- & --
& 0.430 & 0.395 & 0.801 & 0.355 \\

3 & + activity router
& \checkmark & \checkmark & \checkmark & -- & -- & --
& 0.442 & 0.400 & 0.785 & 0.403 \\

4 & + aspect experts
& \checkmark & \checkmark & \checkmark & \checkmark & -- & --
& 0.498 & 0.459 & 0.727 & 0.415 \\

5 & + PDM (w/o CORAL)
& \checkmark & \checkmark & \checkmark & \checkmark & \checkmark & --
& 0.529 & 0.504 & 0.681 & 0.460 \\

\rowcolor{gray!18}
6 & + PDM + CORAL (\textbf{SkillLens})
& \checkmark & \checkmark & \checkmark & \checkmark & \checkmark & \checkmark
& \textbf{0.542} & \textbf{0.517} & \textbf{0.654} & \textbf{0.468} \\

\midrule
& w/o gaze from Full
& \checkmark & -- & \checkmark & \checkmark & \checkmark & \checkmark
& 0.490 & 0.463 & 0.676 & 0.466 \\
\bottomrule
\end{tabular}%
}

\vspace{2pt}
\footnotesize
\emph{Notation.} V: ego/exo video, G: gaze, Rtr: activity router, Asp: aspect experts, PDM: Prototype Dynamics Margin, Sp: Spearman correlation.
\end{table}

\noindent\textbf{Per-level and per-activity behavior.}
Table~\ref{tab:performance_comparison} shows that the proposed model improves both F1 and precision across all four proficiency levels.

Table~\ref{tab:activity_acc_single} shows that each component has
activity-dependent strengths rather than uniformly improving every activity.
Gaze, PDM, router, aspect experts, and CORAL each help different activities,
and some ablated variants outperform the full model in specific cases.
However, no single variant provides the best balance across all activities.
The main finding is therefore that proficiency estimation benefits from
combining complementary evidence: global ego-exo-gaze context, localized PDM
signals, and ordinal calibration.

\begin{table}[t]
\centering
\caption{PDM robustness and control analysis.}
\label{tab:pdm_controls}
\resizebox{\linewidth}{!}{
\begin{tabular}{lcccc}
\toprule
Setting & Acc & MAE & Acc. gain & MAE red. \\
\midrule
Main baseline & 0.4976 & 0.7268 & -- & -- \\
\midrule
\multicolumn{5}{l}{\textit{Action-timestamp dependency}} \\
Proposed w/ action timestamps & 0.5420 & 0.6540 & +4.4 pp & 10.0\% \\
Proposed w/o action timestamps & \textbf{0.5537} & \textbf{0.6114} & \textbf{+5.6 pp} & \textbf{15.9\%} \\
\midrule
\multicolumn{5}{l}{\textit{High-low prototype validity}} \\
\textbf{SkillLens} & \textbf{0.5420} & \textbf{0.6540} & \textbf{+4.4 pp} & \textbf{10.0\%} \\
w/ shuffled proficiency labels & 0.5008 & 0.6837 & +0.3 pp & 5.9\% \\
w/ random prototypes & 0.4976 & 0.7309 & +0.0 pp & -0.6\% \\
w/ activity-average score & 0.5098 & 0.7081 & +1.2 pp & 2.6\% \\
\bottomrule
\end{tabular}
}
\vspace{-1.5em}
\end{table}

\noindent\textbf{Component ablation.}
Table~\ref{tab:ablation_ladder} decomposes the proposed model from a video-only baseline to the full system under a unified evaluation setting.
The activity router and aspect experts strengthen the global backbone, while PDM adds localized high-low prototype evidence from dense action windows.
The CORAL head further calibrates the prediction under the ordered proficiency structure.
Overall, the full system improves accuracy by +11.2 percentage points over the video-only baseline and reduces MAE from 0.796 to 0.654.
The leave-one-out gaze ablation suggests that gaze contributes most clearly when combined with routing, aspect experts, PDM, and ordinal calibration.

\noindent\textbf{PDM robustness and controls.}
Table~\ref{tab:pdm_controls} evaluates whether PDM depends on action timestamps
and whether its high-low prototype signal captures meaningful proficiency
structure.
Replacing action-annotated windows with uniform 2s temporal proposals preserves
the gain over the main baseline, suggesting that PDM can recover high-low
evidence without manually annotated action boundaries at inference.
We further test three controls: shuffled proficiency labels before prototype
construction, random prototype references, and an activity-average score.
The activity-average control replaces action-specific high-low comparisons
with a single activity-level score, removing the ability to compare each local
action against its corresponding high/low prototypes.
The gains largely collapse under these controls, indicating that PDM improves
performance through proficiency-aware and action-specific high-low evidence
rather than simply adding an auxiliary score.

\noindent\textbf{Computational cost.}
Our method contains 5.58M parameters and adds only a post-hoc CORAL calibration head on top of frozen baseline probabilities and two ego/exo PDM auxiliary features.

\begin{figure}[t!]
\centering

\begin{subfigure}{\columnwidth}
    \centering
    
    \includegraphics[width=\linewidth]{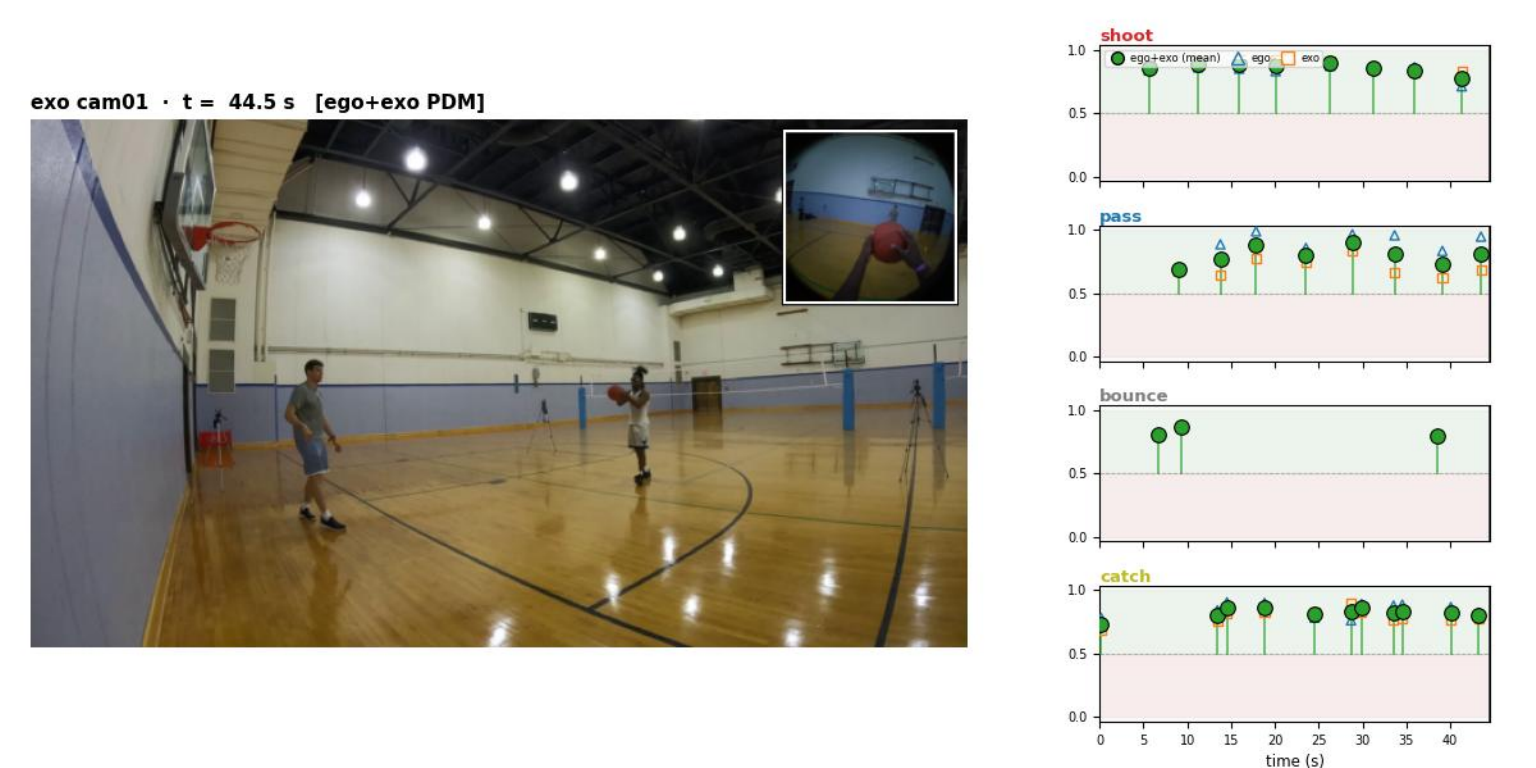}
    \caption{High-proficiency (Late Expert) example on Basketball.}
    \label{fig:qual_a}
\end{subfigure}

\vspace{0.4em}

\begin{subfigure}{\columnwidth}
    \centering
    \includegraphics[width=\linewidth]{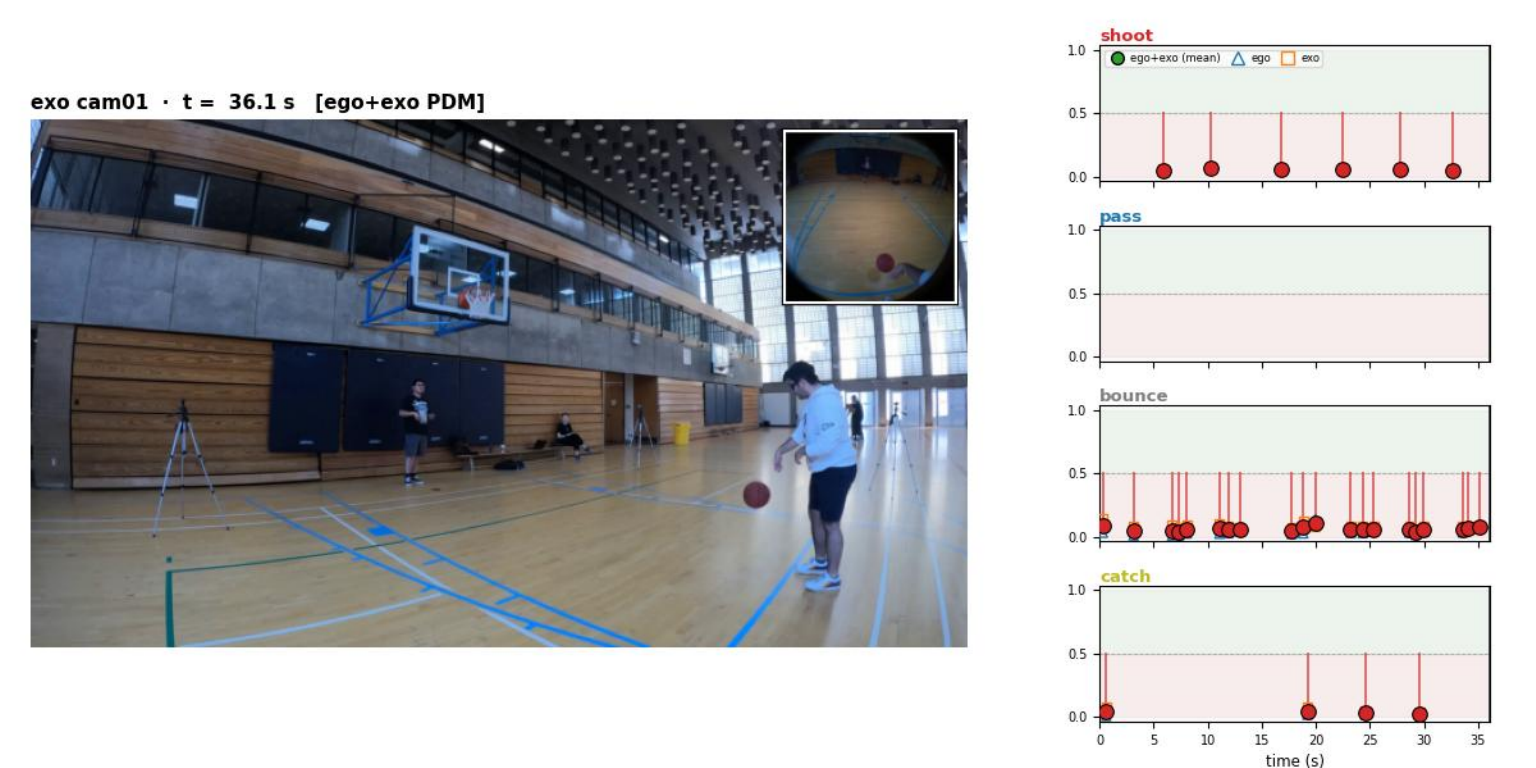}
    \caption{Low-proficiency (Novice) example on Basketball.}
    \label{fig:qual_b}
\end{subfigure}

\caption{Qualitative visualization of PDM expert-likeness scores over atomic-action windows.}
\label{fig:qual}
\vspace{-1.5em}
\end{figure}

\noindent\textbf{Qualitative visualization.}
Fig.~\ref{fig:qual} visualizes per-action PDM expert-likeness scores
over time. Each atomic-action window is encoded with dense MAWS-CLIP-2B
features, compared against train-only high/low prototypes from the same
activity-action bucket, and converted into a percentile score in $[0,1]$.
Scores above $0.5$ indicate more expert-like dynamics, while scores below
$0.5$ indicate more novice-like dynamics. Green circles denote the visualization
mean of ego/exo scores, blue triangles denote ego-only scores, and orange
squares denote exo-only scores. Compared with the novice example, the
higher-proficiency example shows more above-median scores, illustrating that
PDM captures execution-quality differences. \\

\section{Discussion and Limitations}
\begin{table}[t]
\centering
\caption{Activity-action bucket statistics used for prototype construction. \textbf{Total} is the number of training windows; Inter.\ denotes Intermediate.}
\label{tab:supp_bucket_stats}
\setlength{\tabcolsep}{3pt}
\resizebox{\columnwidth}{!}{%
\begin{tabular}{llrrrrr}
\toprule
\textbf{Activity} & \textbf{Action} & \textbf{Total} & \textbf{Late} & \textbf{Inter.} & \textbf{Early} & \textbf{Novice} \\
\midrule
Basketball & shoot & 2,809 & 1,256 & 866 & 544 & 143 \\
Basketball & pass & 761 & 388 & 172 & 160 & 41 \\
Basketball & bounce & 3,096 & 1,040 & 1,203 & 664 & 189 \\
Basketball & catch & 3,600 & 1,609 & 1,123 & 727 & 141 \\
Cooking & pick & 1,012 & 0 & 241 & 377 & 394 \\
Cooking & place & 652 & 0 & 179 & 174 & 299 \\
Cooking & cut & 324 & 0 & 63 & 143 & 118 \\
Cooking & pour\_add & 285 & 0 & 101 & 85 & 99 \\
Cooking & stir\_mix & 67 & 0 & 16 & 31 & 20 \\
Dance & step & 5,533 & 1,417 & 1,044 & 2,592 & 480 \\
Dance & turn\_spin & 2,306 & 666 & 449 & 1,038 & 153 \\
Dance & extend & 1,261 & 223 & 234 & 697 & 107 \\
Dance & swing & 677 & 147 & 94 & 414 & 22 \\
Music & play & 983 & 549 & 270 & 164 & 0 \\
Rock Climbing & grab & 2,453 & 363 & 982 & 488 & 620 \\
Rock Climbing & place\_foot & 1,924 & 284 & 728 & 398 & 514 \\
Soccer & kick & 545 & 397 & 148 & 0 & 0 \\
Soccer & stop & 139 & 90 & 49 & 0 & 0 \\
Soccer & juggle & 290 & 185 & 105 & 0 & 0 \\
Soccer & shoot & 137 & 89 & 48 & 0 & 0 \\
Soccer & pass & 51 & 27 & 24 & 0 & 0 \\
\midrule
\textbf{Total} & -- & \textbf{28,905} & \textbf{8,730} & \textbf{8,139} & \textbf{8,696} & \textbf{3,340} \\
\bottomrule
\end{tabular}%
}
\vspace{-1.3em}
\end{table}

\subsection{Prototype Construction and Bucket Statistics}
Table~\ref{tab:supp_bucket_stats} reports the number of train windows used for
activity-action prototype construction. Each bucket is defined by an activity
and an atomic action type, such as \textit{Basketball:shoot}. Since prototypes
are constructed from window-level action-dynamics features, the counts are
reported at the train-window level.

Prototype construction is performed within each $(\mathrm{activity}, \mathrm{action})$
bucket and separately for each source. Buckets with fewer than five train
windows are skipped. For the high-proficiency side, we first use Late Expert
windows; if fewer than 30 windows are available, we broaden the set to
Late+Intermediate Expert windows. For the low-proficiency side, we first use
Novice windows; if fewer than 20 windows are available, we broaden the set to
Novice+Early Expert windows. If the broadened set still contains fewer than
five windows, no prototype is constructed for that side, and windows that require
missing prototype are excluded from the PDM margin aggregation.

\subsection{Imbalance Across Proficiency Levels}
The table shows that several buckets are imbalanced across proficiency levels.
For example, all Soccer buckets contain no Early Expert or Novice windows, so
their low-proficiency prototypes cannot be constructed even after broadening the
low side. Importantly, we do not replace missing prototypes with coarser
activity-level or cross-action prototypes. This design avoids constructing
prototypes from empty or severely imbalanced buckets while preserving the
activity-action-specific nature of the high-low comparison.

These statistics motivate the use of activity-level percentile normalization
and the prototype control analysis in Table~\ref{tab:pdm_controls}, which test whether the PDM
signal captures a proficiency-aware high-low structure rather than merely
reflecting activity-level or random reference effects.
\begin{figure}[t]
\centering
\includegraphics[width=\columnwidth]{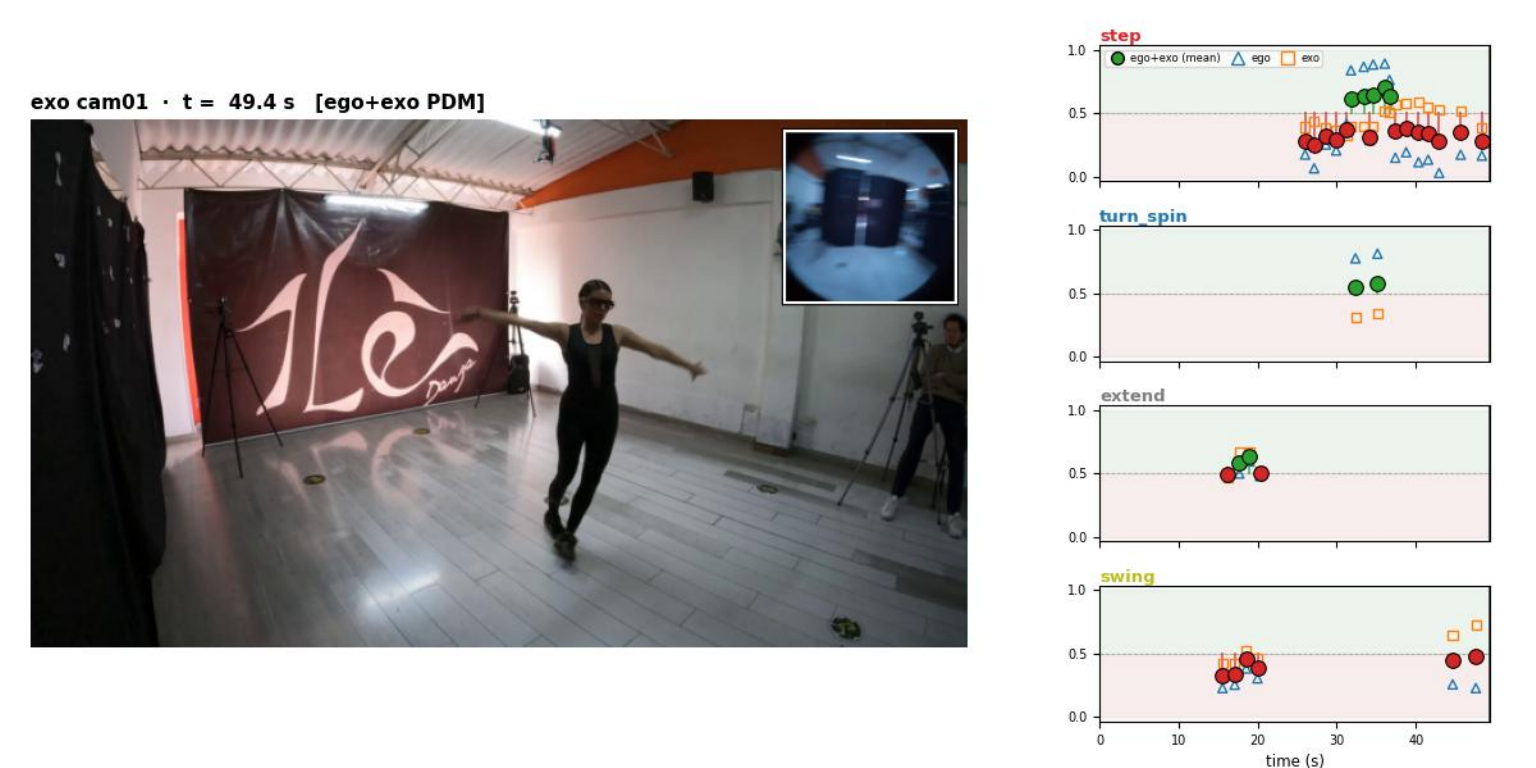}
\caption{
Failure case of action-keyword-based PDM.
}
\label{fig:failure_case}
\vspace{-1.5em}
\end{figure}

\subsection{Complementary Evidence Across Activities}
Our results suggest that proficiency estimation benefits from combining
complementary evidence rather than relying on a single dominant cue. The
per-activity analysis shows that different components are useful for different
activities: gaze, global ego-exo context, localized PDM evidence, and ordinal
calibration each contribute under different motion and attention patterns. This
supports our design choice of integrating global multimodal evidence with
action-level high-low execution cues.

\subsection{Limitations}
Figure~\ref{fig:failure_case} shows a failure case of action-keyword-based PDM.
Although atomic-action descriptions help group local windows and improve over the baseline, broad keywords such as \textit{step} provide ambiguous criteria for proficient execution.
By contrast, dynamics-centered actions such as \textit{turn\_spin} and \textit{swing} offer clearer high-low execution cues.
This suggests that pose or kinematic cues could complement action-keyword-based prototypes for finer-grained skill assessment.

Prototype construction is also affected by class imbalance across
activity-action buckets. We avoid constructing prototypes from empty or
severely underpopulated buckets by broadening proficiency bands within the same
bucket and excluding windows when a valid high or low prototype is unavailable.
However, this can reduce coverage for activities with missing proficiency
levels, such as Soccer. Future work could explore uncertainty-aware or
hierarchical prototype estimation to improve robustness under sparse bucket statistics.
Finally, our current gaze representation provides useful cues but may not fully
capture rapid target switches and fine-grained gaze-action alignment in highly
dynamic activities.

\section{Conclusion}

This paper shows that proficiency estimation is not only a problem of
recognizing what action is performed, but also of identifying where execution
quality is expressed and how it differs across ordered skill levels. The proposed framework (SkillLens) integrates activity-aware ego-exo-gaze evidence, dense
action-window PDM, and CORAL-based ordinal calibration to combine global context
with local high-low execution cues. Our results suggest that balanced
integration of complementary evidence is more important than relying on any
single modality or component.
\bibliographystyle{IEEEtran}
\bibliography{main} 

\end{document}